\documentclass[runningheads,a4paper]{llncs}

\usepackage{amssymb}
\usepackage{graphicx}

\usepackage{makeidx}
\usepackage{graphicx}
\usepackage{amsmath,amssymb} 
\usepackage{color}

\usepackage[linesnumbered,ruled]{algorithm2e}

\usepackage{url}  
\newcommand{\keywords}[1]{\par\addvspace\baselineskip
\noindent\keywordname\enspace\ignorespaces#1}

\begin{document}

\mainmatter  

\title{Wavelet based edge feature enhancement for convolutional neural networks}

\titlerunning{Wavelet based edge feature enhancement for convolutional neural networks}

%
%
\author{D. D. N. De Silva \and S. Fernando\and  I. T. S. Piyatilake\and A. V. S. Karunarathne\\}
\authorrunning{De Silva et al.}

\institute{University of Moratuwa,\\
Bandaranayaka Mawatha, Katubedda, 10400, Sri Lanka}

%
%

\maketitle

\begin{abstract}
Convolutional neural networks are able to perform a hierarchical learning process starting with local features. However, a limited attention is paid to enhancing such elementary level features like edges. We propose and evaluate two wavelet-based edge feature enhancement methods to preprocess the input images to convolutional neural networks. The first method develops feature enhanced representations by decomposing the input images using wavelet transform and limited reconstructing subsequently. The second method develops such feature enhanced inputs to the network using local modulus maxima of wavelet coefficients. For each method, we have developed a new preprocessing layer by implementing each purposed method and have appended to the network architecture. Our empirical evaluations demonstrate that the proposed methods are outperforming the baselines and previously published work with significant accuracy gains.
\end{abstract}

\section{Introduction}
\label{sec:introduction}
		The class of convolution neural networks (CNNs)~\cite{cite_11} is one of the key deep learning architecture and currently a very active field of study. The basic CNNs architecture~\cite{cite_12} comprise alternatively stacked multiple layers that can perform hierarchical feature learning. Over the past decade, CNNs have achieved a huge success compared to conventional machine learning techniques in solving a wide range of problems in different fields of applications such as classification \cite{cite_11,cite_1,cite_28}, object detection \cite{cite_2,cite_3} and recognition~\cite{cite_4}, semantic segmentation~\cite{cite_5} and many others \cite{cite_30,cite_38}. The hierarchical feature learning process~\cite{cite_6} of CNNs are starting with the inputs that are fed into the network and output a class label scores at the end. Initial convolution layers learn basic level features which are local patterns, directly from raw input image pixels by convolving with learnable filter kernels. Following layers of the network develop high level abstractions from extracted features in hierarchical order. Generally, these preliminary level features are edges, colors, and textures. The edges are defined as sharp variations or discontinuities in pixel values. These sharp variation points localize the edge contours which are essential for developing high-level features. 
		
		This makes the edge feature identification is an important and fundamental step of feature learning process. There are many different types of edges exists, created by object boundaries, highlights, shadows, textures, occlusion and etc. Though the convolutions extract the edge features during the preliminary learning, enhancing edge features at the input level can provide an additional boost. We propose two edge enhancing mechanisms using wavelet transform and develop an additional preprocessing layer to implement each method. Then we empirically investigate the performance of proposed methods. The experimental setup uses common and publicly available datasets MNIST, SVHN, and CIFAR-10. Our results demonstrate that both the proposed methods have outperformed the previous work and baselines.
		
		The rest of this paper is organized as follows; We first review the related work in section~\ref{related_work} and discuss our approach for proposed methods including the mathematical background of the wavelet based modulus maxima edge detection. The proposed edge enhancement methods present in section~\ref{approach}. Section~\ref{experimental_setup} describes the implementations and the datasets we have used. Discussion and the obtained results are in section~\ref{results}. 
		
\section{Related Work}
\label{related_work}
	
	The wavelet transform~\cite{cite_15} is capable of providing powerful time-frequency representations. Furthermore, in terms of images, the discrete wavelet transform has the ability to decompose an image into different frequency information levels. There have been several attempts to use this technique in the context of machine learning and in the class of deep learning as follows. Wavelet network in~\cite{cite_32} is a result of a combination between wavelet and neural network \cite{cite_33,cite_34} which is an auto-encoder and constituted with three layers. The authors have examined the effect on classification tasks using 3 different wavelet basis functions. Another work on handwritten digit recognition using SVM and KNN classifiers in~\cite{cite_35} is powered by the wavelet.
	
	In terms of CNNs, the wavelet transform has been used to solve computer vision problems because of its ability to extract diverse frequency information of images. Classification tasks including images~\cite{cite_25,cite_24}, textures~\cite{cite_26}, and multi-scale face super-resolution~\cite{cite_27} are few of them. The work in~\cite{cite_24} pre-processes the input data in the wavelet domain and then fed to the network. They have proposed two methods depending on how the decomposed wavelet coefficients are fused together. Another work done in~\cite{cite_25}, applies the discrete wavelet transform to extract features and a neural network used to perform classification by the resultant feature vector. Authors in~\cite{cite_26} have employed discrete wavelet transform to convert images into the wavelet domain and exploit important features to perform classification using a CNN. Furthermore, Liu \textit{et al.}~\cite{cite_37} introduced a multi-level wavelet CNN model for image restoration. However, most of the work paid attention to decomposing wavelet coefficients directly rather than processing them to enhance particular features like edges.
	
	Therefore, we introduce two mechanisms to apply the wavelet transform to enhance the edge features in input images to improve the classification of CNNs. The concept of edge detection~\cite{cite_14,cite_7,cite_8,cite_13} is referred to identifying edges starting from contours of small structures to boundaries of large visual objects in an image. Mallat~\cite{cite_14} presented a multi-scale edge detection algorithm using the wavelet transform that develops edge representations by finding the local maxima of a wavelet transform modulus which is also equivalent to gradient-based edge detection. This will be the base technique for one of our purposed method. The other method is a naive edge detection mechanism also based on wavelet transformation with limited coefficients reconstruction after decomposition.

\section{Approach}
\label{approach}
	
\subsection{Discrete wavelet transform}
\label{dwt}
	
\begin{figure*}[h]
		\begin{center}
			\includegraphics[width=\linewidth]{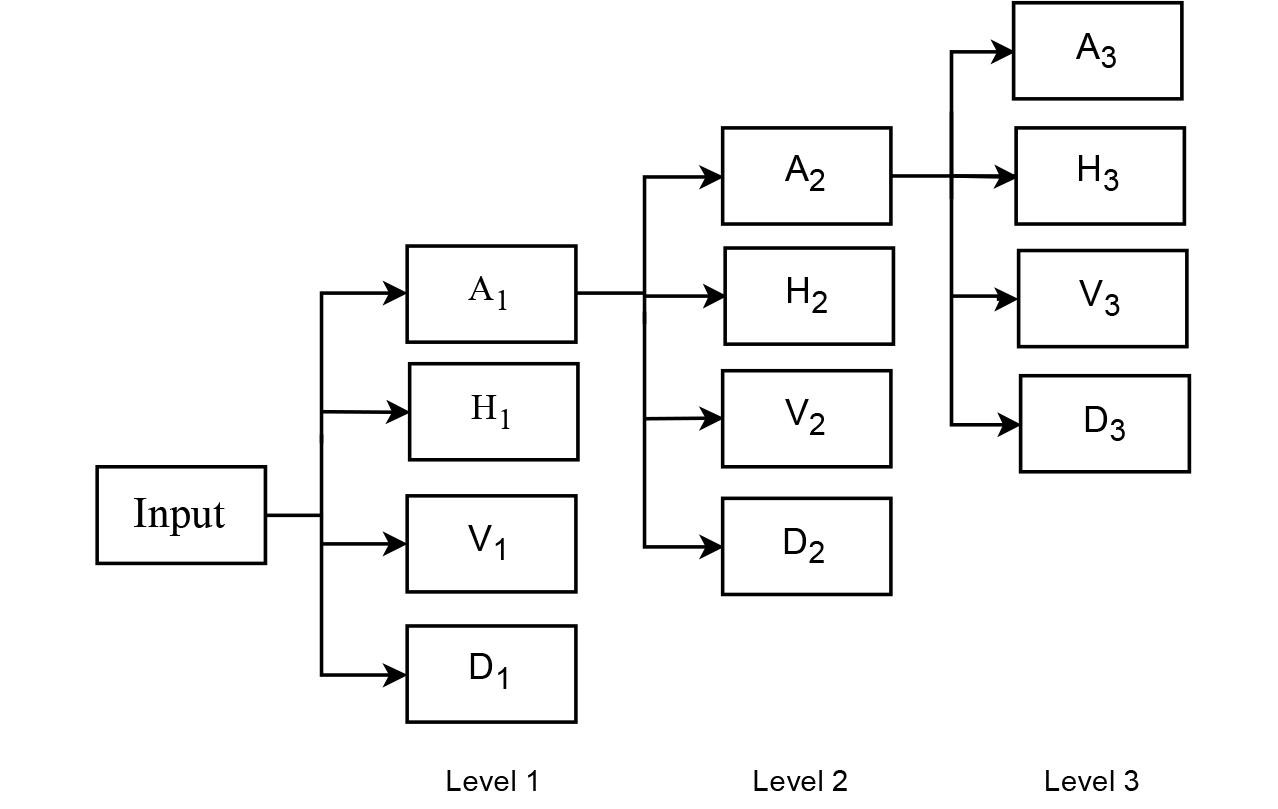}
		\end{center}
		\caption{Discrete wavelet transform decompositions.}
		\label{fig.1}
	\end{figure*}

	The discrete wavelet transform~\cite{cite_17,cite_16} is heavily used in image processing applications~\cite{cite_19,cite_18}. It decomposes an image into different levels of frequency interpretations by simultaneously passing through a set of low pass and high pass filters. Resulting wavelet coefficients contain decomposed image detail information and are known as wavelet coefficients. There are two main types of coefficients: detail coefficients and approximation coefficients. One level of decomposition down samples the coefficients by a factor of two to prevent information redundancy. After the first level of decomposition, resultant approximation coefficient can be subjected to further wavelet transform decompositions to generate second level coefficients and so on as illustrated in Fig. \ref{fig.1}. The detail coefficients contain the high-frequency information and are generated by passing through the high pass filters. The approximation coefficients provide less detailed low-frequency versions of the original image that are generated by passing through the low pass filters. Since an image is a two dimensional (2D) signal or data, the discrete wavelet transform is applied in a 2D manner. This 2D transform is applied to an image as two operations of one-dimensional discrete wavelet transform along the rows and columns separately which results in four different wavelet coefficients. Resultant coefficients comprise three detail coefficients containing vertical, horizontal and diagonal details, and one approximation coefficient containing low-frequency details.
	
\subsection{Mathematical background for modulus maxima edge detection}
\label{math_background}
	
	Basic gradient-based edge detection process analyzes the image points from their first or second order derivatives to detect sharp variation points where edges are localized. Extrema of the first derivative and zero crossings of the second derivative are corresponding to sharp variation points where edges have occurred. Mallat~\cite{cite_14} method establishes a systematic relationship between the wavelet transform and the edge detection which is the base mechanism of our proposed modulus maxima method. It describes that the wavelet transform of each image point is referring to the derivative at a given scale. This can prove by relating the wavelet with differentiable smoothing function whose integral is equal to $1$. One can easily choose Gaussian as the differentiable smoothing function. Since this study is interested in image data, computations are performed two-dimensionally along the image axis $x$ and $y$. There defined $\psi^x(x,y)$ and $\psi^y(x,y)$ as the first derivative of differentiable smoothing function  $\theta(x,y)$ whose integral over $x$ and $y$ is equal to $1$ and converges to $0$ at the infinity. Considering in multiscale, scaling factor $s$ is added as the scale dilation.
	
	\begin{equation}\label{eq.1}
	\psi_{s}^x(x,y)=\frac{\partial \theta(x,y)}{\partial x}\hspace{0.5cm}\mathrm{and}\hspace{0.5cm}\psi_{s}^y(x,y)=\frac{\partial \theta(x,y)}{\partial y}.
	\end{equation}
	The wavelet transform is computed by convolving an image $f(x,y)$ with a dilated wavelet $\psi_{s}(x,y)$ and is given by,
	\begin{equation}\label{eq.2}
	W_{s}f(x,y)=f*\psi_{s}(x,y).
	\end{equation}
	As we are processing discrete images, the discrete wavelet transform is considered. Therefore, the dilation can impose from continues scaling factor to dyadic sequence $(2^{j})_{j\in\mathbb{Z}}$ that refers to $2D$ dyadic wavelet transform. The $2D$ smoothing function is $\theta_{2^{j}}(x,y)$. The image function $f(x,y)$ convolves with the smoothing function and then,
	\begin{equation}\label{eq.3}
	\psi_{2^{j}}^x(x,y)=\frac{\partial {\theta_{2^{j}}}(x,y)}{\partial x}\hspace{0.5cm}\mathrm{and}\hspace{0.5cm}\psi_{2^{j}}^y(x,y)=\frac{\partial {\theta_{2^{j}}}(x,y)}{\partial y}.
	\end{equation}
	\begin{equation}\label{eq.4}
	W_{2^{j}}^xf(x,y)=f*\psi_{2^{j}}^x(x,y)\hspace{0.5cm}\mathrm{and}\hspace{0.5cm}W_{2^{j}}^yf(x,y)=f*\psi_{2^{j}}^y(x,y).
	\end{equation}
	Then the gradient vector is given by,
	\begin{equation}\label{eq.5}
	\left(\begin{array}{c}{W_{2^{j}}^xf(x,y)}\\ {W_{2^{j}}^yf(x,y)}\end{array}\right)=2^{j}\left(\begin{array}{c}{\frac{\partial }{\partial x}(f*{\theta_{2^{j}}})(x,y)}\\ {\frac{\partial }{\partial y}(f*{\theta_{2^{j}}})(x,y)}\end{array}\right)=2^{j}\nabla(f*\theta_{2^{j}})(x,y).
	\end{equation}
	Local minima or maxima of the wavelet transform (first-order derivative) corresponds to variations of the pixel intensities. The absolute value of the first derivatives can either be a maximum or a minimum where local maxima correspond to sharp variations and local minima correspond to gradually varyings. Both the gradient vector and the direction vector can compute at each image point;
	\begin{equation}\label{eq.6}
	M_{2^{j}}f(x,y)=\sqrt{|W_{2^{j}}^x(x,y)|^2+|W_{2^{j}}^y(x,y)|^2}.
	\end{equation}
	Direction of the gradient with the horizontal axis is given by,
	\begin{equation}\label{eq.7}
	A_{2^{j}}f(x,y)=\arctan\left(\frac{W_{2^{j}}^yf(x,y)}{W_{2^{j}}^xf(x,y)}\right).
	\end{equation}
	
\begin{figure*}[h]
		\begin{center}
			\includegraphics[width=\linewidth]{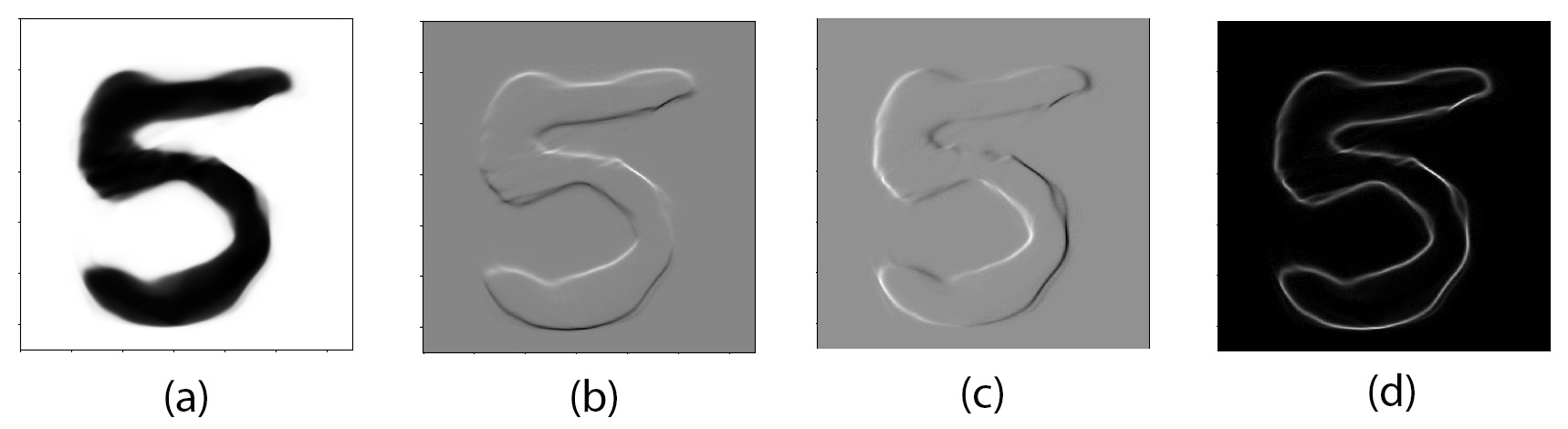}
		\end{center}
		\caption{Graphical representation of modulus maxima process (sample image from MNIST dataset). (a): Original image; (b): horizontal detail coefficients obtained from wavelet transform; (c): vertical detail coefficients obtained from wavelet transform; (d): Edge representation development from local modulus maxima.}
		\label{fig.2}
	\end{figure*}

	Sharp variation point locations are given by the local maxima of $M_{2^{j}}f(x,y)$ and the directions are given by $A_{2^{j}}f(x,y)$. The modulus has a local maximum in direction of the gradient corresponding to the locations where sharp intensity variation points exist. In order to find these local maxima of the modulus at each image point, modulus of the gradient is compared with its local neighborhood. Positions of the local modulus maxima are corresponding to the image edges along with their directions. Fig. \ref{fig.2} graphically illustrates the outputs of steps of the modulus maxima process.
	
\section{Methodology}
\label{methodology}
	
	This study proposes two methods to develop edge feature enhanced input images to CNNs. Firstly a naive edge enhancement method is proposed and the modulus maxima method is the purposed second as explained from the section \ref{math_background}.
	
\subsection{Naive method}
\label{naive_method}
	
\begin{figure*}[h]
		\begin{center}
			\includegraphics[width=\linewidth]{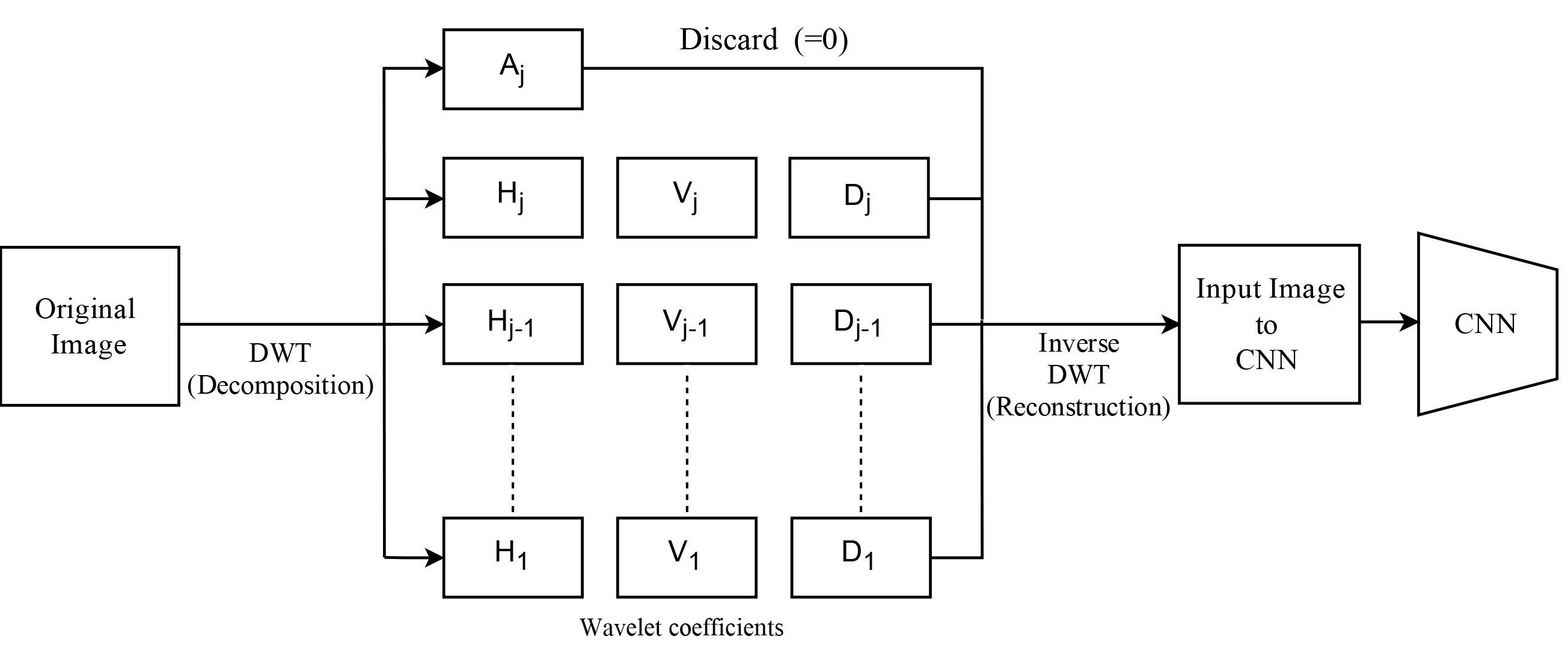}
		\end{center}
		\caption{Naive edge enhancement method.}
		\label{fig.3}
	\end{figure*}

	The naive method proposes a primary mechanism to enhance edge features of input images to CNNs using the wavelet transform.  As illustrated in figure \ref{fig.3}, this process starts by applying the discrete wavelet transform to decompose an input into series of wavelet coefficients. The coarsest approximation coefficient and several detail coefficients are generated depending on the number of decomposition levels using `Haar' wavelet as basis wavelet function. Haar wavelet~\cite{cite_23} is the simplest wavelet basis function that has squared shape and commonly used in image processing applications. secondly, the resulting coarsest approximation coefficient is discarded in order to remove lowest frequency details. Then the input image is reconstructed back by inverse wavelet transform using the same wavelet basis function with remaining detail coefficients. The reconstructed image is fed as the input to the CNN. The fed input is a basic level edge enhanced representation of the original image.
	
\subsection{Modulus maxima method}
\label{modulus_maxima_method}
	
\begin{figure*}[h]
		\begin{center}
			\includegraphics[scale=0.38]{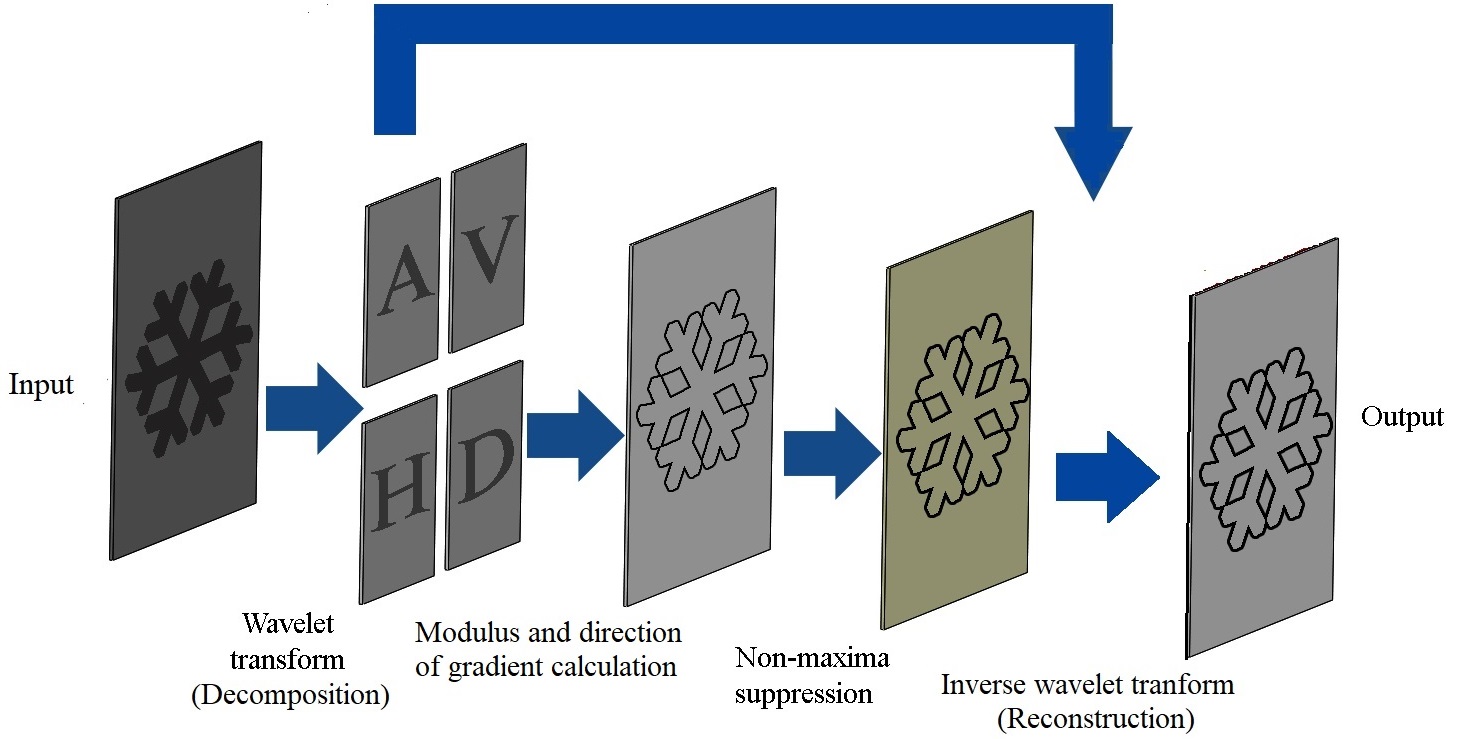}
		\end{center}
		\caption{Graphical illustration of modulus maxima process.}
		\label{fig.4}
	\end{figure*}

	The modulus maxima methodology from section \ref{math_background} is the base mechanism for the second proposed method. It is able to develop edge feature enhanced representations out of original input images. This process is initiating by Gaussian smoothing of the input image for noise reduction. The wavelet transform of the smoothed input image is then performed using `Haar' wavelet. Since the proposed method is interested in preserving most salient edge information, only the first level of the wavelet transform is performed. As the discrete wavelet transform decomposes the input image into four different coefficients as explained in section \ref{dwt}, Only the detail coefficients which are containing high-frequency information including horizontal, vertical, and diagonal coefficients are being taken into account from this point onwards. The modulus of the detail coefficients are then calculated at each image point according to (\ref{eq.6}) and the directions are calculated using (\ref{eq.7}). The edge representations are developed by finding the local maximal of the modulus along the gradient directions subsequently. This is done by applying non-maximal suppression over the local neighborhoods. The non-maximal suppression is explained in Alg. \ref{alg.1}. As a post-processing, a proper thresholding is applied to improve the developed map. The final output of the process is obtained by reconstructing the image back by inverse wavelet transform from the built edge representation and the approximation coefficient remaining from the wavelet transform of the original input. Fig. \ref{fig.4} graphically illustrates the described process. The output of the reconstruction becomes the input to CNN for feature learning and classification.
	
		\begin{algorithm}[h]
		\KwIn{Modulus $(M)$ and angle $(A)$ of the gradient vector computed by (\ref{eq.6}) and (\ref{eq.7}) respectively} 
		\KwOut{Local maxima $(LM)$ of the modulus}
		\For(\tcp*[f]{$i,j$ are image indices}){$i$}{
			\For{$j$}{
				
				\If{$A_{i,j}$ in between $\left[-\frac{\pi}{8},\frac{\pi}{8}\right]$ and$ \left[\pi-\frac{\pi}{8}, \pi+\frac{\pi}{8}\right]$}{\If{$M_{i,j}>M_{i+1,j}$ and $M_{i,j}>M_{i-1,j}$}{$LM_{i,j}\gets M_{i,j}$}} 
				
				\ElseIf{$A_{i,j}$ in between $\left[\frac{\pi}{2}-\frac{\pi}{8},\frac{\pi}{2}+\frac{3\pi}{8}\right]$ and$ \left[\frac{3\pi}{2}-\frac{\pi}{8}, \frac{3\pi}{2}+\frac{\pi}{8}\right]$}{\If{$M_{i,j}>M_{i,j+1}$ and $M_{i,j}>M_{i,j-1}$}{$LM_{i,j}\gets M_{i,j}$}}
				
				\ElseIf{$A_{i,j}$ in between $\left[\frac{\pi}{4}-\frac{\pi}{8},\frac{\pi}{4}+\frac{3\pi}{8}\right]$ and$ \left[\frac{5\pi}{4}-\frac{\pi}{8}, \frac{5\pi}{4}+\frac{\pi}{8}\right]$}{\If{$M_{i,j}>M_{i+1,j+1}$ and $M_{i,j}>M_{i-1,j-1}$}{$LM_{i,j}\gets M_{i,j}$}}
				
				\Else{\If{$M_{i,j}>M_{i+1,j-1}$ and $M_{i,j}>M_{i-1,j+1}$}{$LM_{i,j}\gets M_{i,j}$}}
			}
		}
		
		\caption{Non-maximal suppression.}
		\label{alg.1}
	\end{algorithm}
	
\section{Experimental Setup}
\label{experimental_setup}
	
\begin{figure*}[h]
		\begin{center}
			\includegraphics[width=\linewidth]{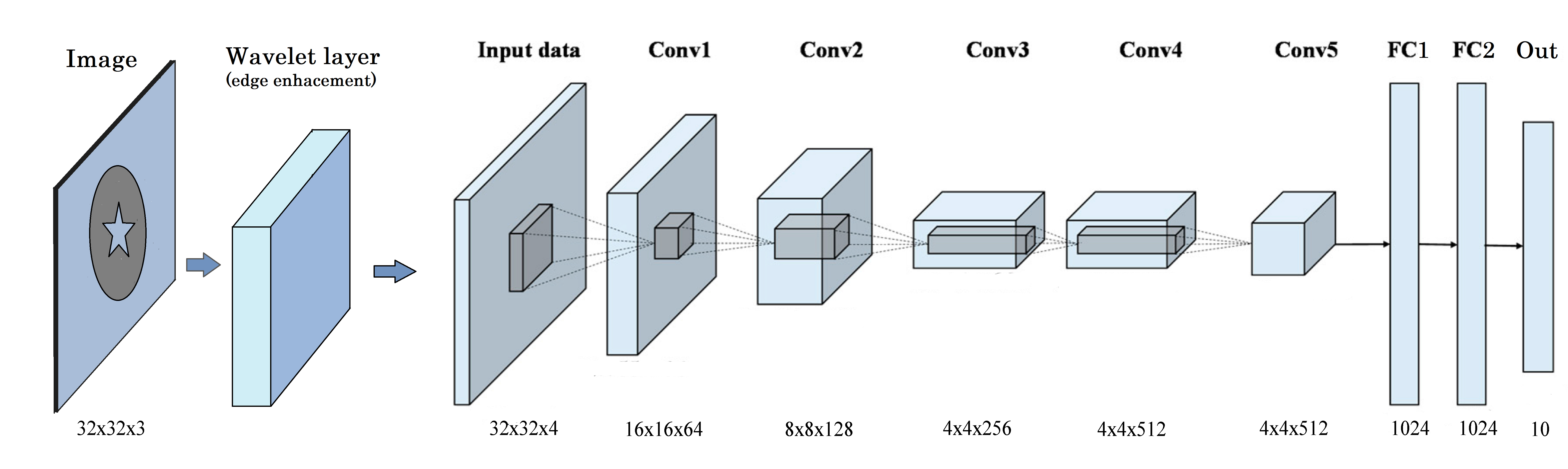}
		\end{center}
		\caption{Overview of the CNN architecture implementation.}
		\label{fig.5}
	\end{figure*}

	In order to evaluate how proposed methods are performing, both methods are implemented to  CNNs, then trained and tested from the scratch. The implementation of the naive edge enhancement method is labeled from now on as NEE-\textit{CNN} and the modulus maxima method is labeled as MMEE-\textit{CNN} for ease of explanation. The annotation \textit{CNN} denotes the base CNN architecture and will be changed depending on the baseline CNN and the compared previous work. For example, if the base network is AlexNet, proposed methods will be labeled as NEE-AlexNet and MMEE-AlexNet. For each method, a new data processing layer is developed by implementing the proposed method and is appended to on top of the CNN architecture as displayed in Fig. \ref{fig.5}. This layer develops edge feature enhanced input images and feed them to the first convolution layer of the CNN. As per investigations, the complete system is then trained and tested on several different datasets.
	
	Firstly, we employed AlexNet~\cite{cite_11} as the base CNN architecture to implement proposed methods. AlexNet is composed of 7 layers including 5 convolutional layers and 2 fully connected layers. In AlexNet architecture, each convolution layer is followed by ReLu activation function to introduce non-linearity. Furthermore, batch normalization~\cite{cite_36} applied version of AlexNet is also employed to use as an additional baseline. We used dropout~\cite{cite_11} with 0.8 of keep probability for fully connected layers while training to prevent over-fitting. The developed preprocessing layer is appended just before the first convolution layer of each network. This layer processes raw input images to feed edge enhanced images into the network for further feature learning. We trained four architectures on three different datasets and the obtained results are discussed in section \ref{results}.  
	
	To investigate how the developed methods perform in classification tasks, all three networks are trained separately on three data sets. The trained networks are tested on the testing data portion of each dataset.  Most common and publicly available datasets are selected: MNIST~\cite{cite_12}, SVHN~\cite{cite_22}, and CIFAR-10~\cite{cite_29}. MNIST is a popular and one of the most common preliminary dataset in machine learning practices. The dataset composed of 28 by 28 gray-scale handwritten digit images from 0-9 with 60,000 full training data and 10,000 full testing data. SVHN dataset is a real-world street view house number images extracted from Google maps street view. It contains 65931 training data and 26032 testing data. CIFAR-10 dataset is composed of 32 by 32 images with 50,000 training examples and 10,000 testing examples under 10 classes of real-world objects like car, airplane, dog, and horse. All datasets have used without any data augmentation.
	
\section{Results}
\label{results}
	
	We trained our both implementations, classic AlexNet, and AlexNet with BN from the scratch on each dataset. Furthermore, we implemented and tested our methods on network models that have been used in previous work~\cite{cite_32,cite_24}. As an overview, the obtained results exhibit that both proposed methods are outperforming the baselines and previous work for all datasets.
	
	Table \ref{results-table-cifar-10-alexnet} shows classification results of CIFAR-10 for network models NEE-AlexNet, MMEE-AlexNet, AlexNet and AlexNet-BN (Batch normalization applied version). Moreover, we compared the proposed methods with the previous work~\cite{cite_24} and results are displayed in Table \ref{results-table-cifar-10-previous_work}. Note that this experiment has employed the same network from compared work for CIFAR-10 dataset. As shown in the Table \ref{results-table-cifar-10-alexnet}, our networks are significantly outperforming both AlexNet baselines. The results are yielding the accuracy of the most accurate baseline, AlexNet-BN by 0.63\% and 0.47\% increment for the modulus maxima method and the naive method respectively. Results comparison with~\cite{cite_24}, also exhibits nearly 1.5\% accuracy gain for the modulus maxima method and 1.2\% gain for the naive method over best classification accuracy obtained. Between our two proposed methods, the modulus maxima method has been able to show the most success.

	\setlength{\tabcolsep}{4pt}
	\begin{table}[h]
		\centering
		\caption{Test accuracies on CIFAR-10 with AlexNet as baseline architecture.}
		\label{results-table-cifar-10-alexnet}
		\begin{tabular}{lc}
			\hline\noalign{\smallskip}
			Network  & Accuracy \%  \\
			\noalign{\smallskip}
			\hline
			\noalign{\smallskip}
			\textbf{MMEE-AlexNet (Proposed)} & \textbf{89.63}    \\
			\textbf{NEE-AlexNet (Proposed)} & \textbf{89.47}      \\
			AlexNet  & 87   	\\
			AlexNet-BN	 & 89     \\
			\hline
		\end{tabular}
	\end{table}
	\setlength{\tabcolsep}{1.4pt}
	
	\setlength{\tabcolsep}{4pt}
	\begin{table}[h]
		\centering
		\caption{Results comparison with~\cite{cite_24} and proposed methods for CIFAR-10.}
		\label{results-table-cifar-10-previous_work}
		\begin{tabular}{lc}
			\hline\noalign{\smallskip}
			Network  & Accuracy \%  \\
			\noalign{\smallskip}
			\hline
			\noalign{\smallskip}
			CNN & 77.53    \\
			CNN-WAV 2 & 76.42    \\
			CNN-WAV 4 & 85.67    \\
			\textbf{MMEE-CNN (Proposed)} & \textbf{87.21}    \\
			\textbf{NEE-CNN (Proposed)} & \textbf{86.84 }    \\
			\hline
		\end{tabular}
	\end{table}
	\setlength{\tabcolsep}{1.4pt}
	
	As we investigated for CIFAR-10, we also employed AlexNet as the baseline network architecture for the MNIST too. The obtained results are displayed in the Table \ref{results-table-MNIST-alexnet}. Even the proposed methods are exceeding the baselines, the gain is not as significant as CIFAR-10. It only showed 0.04\% and 0.07\% accuracy yield from Alexnet-BN baseline for MMEE-AlexNet and NEE-Alexnet respectively. This is because of the simplicity of MNIST and AlexNet easily reach high accuracies without much effort. However, still, the results show that the proposed methods are further improving the classification accuracies. Table \ref{results-table-MNIST-previous-work} compares the results from our methods with~\cite{cite_32}. This previous work has used different wavelet basis functions and they have developed a wavelet network to extract features. However, our methods have outperformed this work by shallow margins.
	
	\setlength{\tabcolsep}{4pt}
	\begin{table}[h]
		\centering
		\caption{Test accuracies on MNIST with AlexNet as baseline architecture.}
		\label{results-table-MNIST-alexnet}
		\begin{tabular}{lc}
			\hline\noalign{\smallskip}
			Network  & Accuracy \%  \\
			\noalign{\smallskip}
			\hline
			\noalign{\smallskip}
			\textbf{MMEE-AlexNet (Proposed)} & \textbf{99.29}    \\
			\textbf{NEE-AlexNet (Proposed)} & \textbf{99.26}      \\
			AlexNet  & 99.17   	\\
			AlexNet-BN	 & 99.22     \\
			\hline
		\end{tabular}
	\end{table}
	\setlength{\tabcolsep}{1.4pt}
	
	\setlength{\tabcolsep}{4pt}
	\begin{table}[ht]
		\centering
		\caption{Results comparison with~\cite{cite_32} and proposed methods for MNIST.}
		\label{results-table-MNIST-previous-work}
		\begin{tabular}{lc}
			\hline\noalign{\smallskip}
			Network  & Accuracy \%  \\
			\noalign{\smallskip}
			\hline
			\noalign{\smallskip}
			\textbf{Modulus maxima edge enhancement approach (Proposed)} & \textbf{99.3}  \\
			\textbf{Naive edge enhancement approach (Proposed)} & \textbf{99.26}      \\
			Wavelet network approach with mexican hat wavelet  &   94.2 	\\
			Wavelet network approach-Morlet wavelet	 & 99.21     \\
			Wavelet network approach-rasp wavelet  & 99.2   	\\
			\hline
		\end{tabular}
	\end{table}
	\setlength{\tabcolsep}{1.4pt}
	
	For further evaluations, experimental results on SVHN dataset are shown in Table \ref{results-table-SVHN-alexnet}. Our method again outperforms the baseline CNN by showing accuracy gains 0.45\% and 0.65\% for the modulus maxima method and the naive method respectively. The significance of the accuracy gain proves that the proposed methods are successfully performing in assisting the feature learning process by providing edge enhanced representations.   
	
	\setlength{\tabcolsep}{4pt}
	\begin{table}[!b]
		\centering
		\caption{Test accuracies on SVHN with AlexNet as baseline architecture.}
		\label{results-table-SVHN-alexnet}
		\begin{tabular}{lc}
			\hline\noalign{\smallskip}
			Network  & Accuracy \%  \\
			\noalign{\smallskip}
			\hline
			\noalign{\smallskip}
			\textbf{MMEE-AlexNet (Proposed)} & \textbf{94.43}    \\
			\textbf{NEE-AlexNet (Proposed)} & \textbf{94.16}      \\
			AlexNet  & 93.24   	\\
			AlexNet-BN	 & 93.81     \\
			\hline
		\end{tabular}
	\end{table}
	\setlength{\tabcolsep}{1.4pt}
	
	Identifying visual objects in images are mostly depend on its shape or in other words, the combinations of edges of visual objects. During the classification, learning of the first level convolution layers mostly relies on these edge features. Hence the improved feature representations provided by the proposed methods are effectively assisting the learning to achieve better classification accuracy. The Obtained results have shown that enhancing edge features can significantly affect the classification accuracy and hence improve the learning process. Furthermore, results are further confirming that the modulus maxima method develops richer feature enhanced representations and performs better than the naive method.   
	
\section{Conclusion}
\label{conclusion}

	We have proposed and empirically evaluated two wavelet based edge enhancement mechanisms to pre-process the input images to convolutional neural networks. The aim of this preprocessing is to improve and assist the learning of the network by enhancing the edge features. The first method performs the process by discarding the coarsest approximation coefficient generated from the discrete wavelet transform of the original input image and then reconstructed by the inverse wavelet transform with the remaining detail coefficients. Secondly, a more complex method is proposed to detect edges by finding local maxima of the modulus of wavelet coefficients as discussed in section \ref{modulus_maxima_method}. The obtained results from the experiments conducted have shown that the proposed methods achieved better classification accuracies compared to the baselines and the previous work. It is notable that the developed systems achieve success in classifying images where the edges are prominent features to be learned during the classification. The Haar wavelet is used as the base wavelet in both proposed methods. There are other wavelets also available in the literature that are also suitable for this application such as Daubechies and Morlet. Thresholding operation of the modulus maxima methodology can implement as a learnable process alongside with the CNN so it can produce better outputs and `\cite{cite_39}.\\

\bibliographystyle{splncs04}
\bibliography{egbib}

\end{document}